\documentclass[twocolumn]{article}

\usepackage[top=0.8in, bottom=0.8in, left=0.8in, right=0.8in]{geometry}
\usepackage{amsfonts, amssymb, amsmath, dsfont}
\usepackage[linktoc=page, colorlinks, linkcolor=blue, citecolor=blue]{hyperref}
\usepackage{enumerate}
\usepackage{multirow}
\usepackage{float}
\usepackage[ruled,boxed,vlined]{algorithm2e}
\numberwithin{equation}{section}
\usepackage[dvipsnames]{xcolor} 
\usepackage{graphicx}
\usepackage{subcaption}
\usepackage{epstopdf}
\usepackage{todonotes}
\usepackage{tikz}
\usetikzlibrary{decorations.shapes}
\usetikzlibrary{decorations.pathreplacing}
\tikzstyle{cnode} = [draw, circle,scale=0.6]
\tikzstyle{level 1} = [level distance=.25\textwidth, sibling distance=.5\textwidth]
\tikzstyle{level 2} = [level distance=.15\textwidth, sibling distance=.25\textwidth]
\tikzstyle{level 3} = [sibling distance=.125\textwidth]

\def \R {\mathbb{R}}

\def \K {\mathcal{K}}

\DeclareMathOperator{\sign}{sign}

\DeclareMathOperator{\argmax}{argmax}
\DeclareMathOperator{\argmin}{argmin}

\begin{document}
	
\title{A Study of Clustering Techniques and Hierarchical Matrix Formats\\
 for Kernel Ridge Regression}

\author{Elizaveta Rebrova\footnote{Department of Mathematics,
University of Michigan, 530 Church St, Ann Arbor, MI 48109,
U.S.A. erebrova@umich.edu} \and
Gustavo Ch{\'a}vez\footnote{Computational Research Division,
Lawrence Berkeley National Laboratory, 1 Cyclotron Road, Berkeley,
CA 94720, U.S.A. \{gichavez,liuyangzhuan,pghysels,xsli\}@lbl.gov}  \and Yang Liu\footnotemark[2] \and Pieter
Ghysels\footnotemark[2] \and Xiaoye Sherry Li\footnotemark[2]}

\date{}
\maketitle

\begin{abstract}

We present memory-efficient and scalable algorithms for kernel
methods used in machine learning. 
Using hierarchical matrix approximations for the kernel matrix the memory requirements,
the number of floating point operations, and the execution time are
drastically reduced compared to standard dense linear algebra
routines. We consider both the general $\mathcal{H}$ matrix hierarchical
format as well as Hierarchically Semi-Separable (HSS) matrices. Furthermore, we
investigate the impact of several preprocessing and clustering
techniques on the hierarchical matrix compression. Effective
clustering of the input leads to a ten-fold increase in efficiency
of the compression. The algorithms are implemented using the	
STRUMPACK solver library. These results confirm that ---
with correct tuning of the hyperparameters --- classification using
kernel ridge regression with the compressed matrix does not lose
prediction accuracy compared to the exact --- not compressed ---
kernel matrix and that our approach can be extended to
$\mathcal{O}(1M)$ datasets, for which computation with the full
kernel matrix becomes prohibitively expensive. We present numerical 
experiments in a distributed memory environment up to 1,024 processors
of the NERSC's Cori supercomputer using well-known datasets 
to the machine learning community that range from dimension 8 up to 784.
\end{abstract}

\vspace{-10px}
\section{Introduction}
	
Kernel methods play an important role in a variety of applications in
scientific computing and machine learning (see, for example,
\cite{kernel methods}). The idea is to implicitly map a set of data to
a high-dimensional feature space via a kernel function, which allows
performing a more sensitive training procedure. Given $n$ data points
$x_1, \ldots, x_n \in \R^d$ and a kernel function
$\K: \R^d \times \R^d \to \R$, the corresponding kernel matrix
$K \in \R^{n\times n}$ is defined as $K_{ij} = \K(x_i, x_j)$. Solving
linear systems with kernel matrices is an algebraic procedure required
by many kernel methods. One of the simplest examples is kernel ridge
regression in which one solves a system with the matrix
$K + \lambda I$, with $\lambda$ a regularization parameter and $I$ the
identity matrix. The limitation of this approach is in the lack of
scalability. The number of data-points, $n$, is typically very large,
and a direct solve would require $\mathcal{O}(n^3)$ operations, even
requiring $\mathcal{O}(n^2)$ complexity just to construct the full
kernel matrix.

Acceleration of kernel methods has been studied extensively in
scientific computing research, e. g.~\cite{GM, Bach, rand kernel}, mostly
using low-rank matrix approximations for $K$. This is, however, based
on an assumption which is not valid in general. Consider for example one of the 
kernel matrices, defined by	the Gaussian radial basis function:
    
\begin{equation}\label{K}
K_{ij} :=  \exp\left( - \frac{1}{2}\frac{\|x_i - x_j\|^2}{h^2}\right).
\end{equation}
    
Note that for $h \to 0$, $K$ approaches the identity matrix, while for
$h \to \infty$ it is nearly a rank one matrix (all elements
$K_{ij} \sim 1$). Intermediate values of $h$ interpolate between these
``easy'' cases.  The value of $h$ is determined by the expected accuracy of
the machine learning algorithm on a certain dataset (e.g. by 	cross-validation).
Therefore, we cannot simply assume the ``easy'' structure of the
matrix.  However, the off-diagonal part of the kernel matrix typically
has a fast singular value decay, which means kernel matrices are good
candidates for hierarchical low-rank solvers like
STRUMPACK~\cite{ghysels2017robust,1,3}.
	
\begin{figure}[H]
\centering
\begin{subfigure}[t]{.23\textwidth}
	\centering
	\includegraphics[height=1.2in]{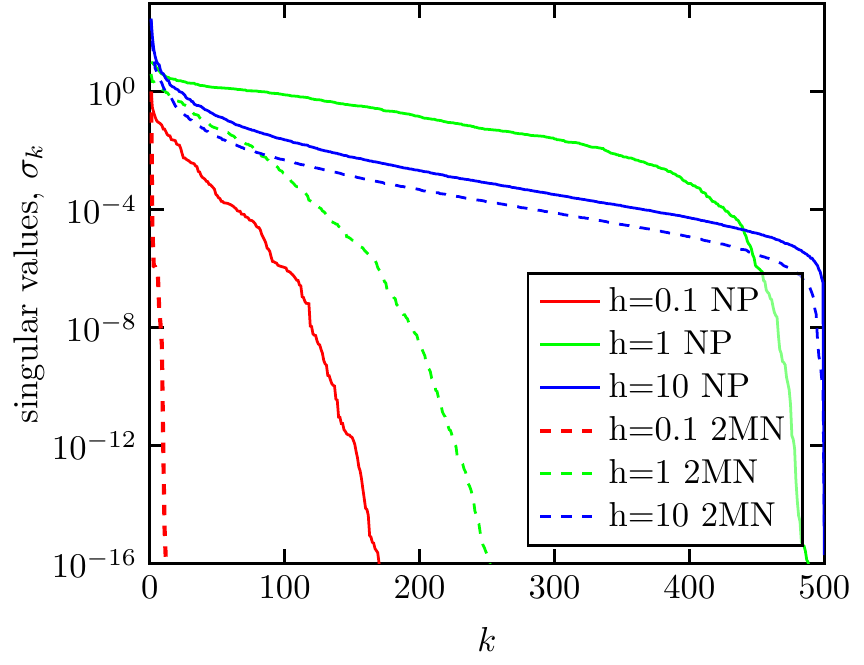}
	\caption{off-diagonal $500 \times 500$ block of the kernel matrix}
	\label{fig:svdecay}
\end{subfigure}%
\hspace{0.01in}
\begin{subfigure}[t]{.23\textwidth}
	\centering
	\includegraphics[height=1.2in]{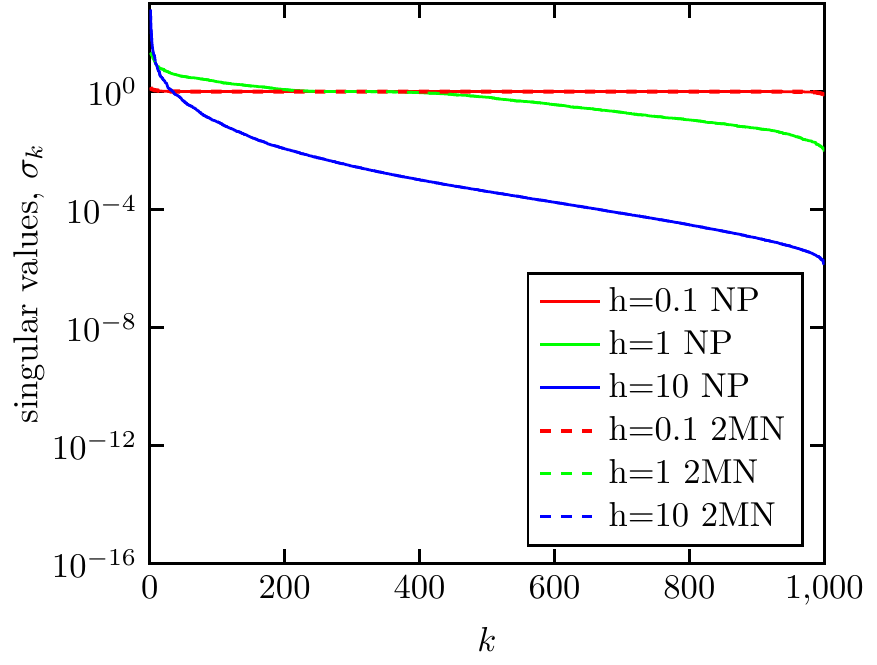}
	\caption{full kernel matrix}
	\label{fig:sv-full}
\end{subfigure}
\caption{GAS1K: singular values with and without two-means (2MN) preprocessing.}
\end{figure}

For a discussion on the existence of efficient far-field compression of the off-diagonal blocks we refer the reader to \cite{march2015} and the references therein.
To empirically confirm the off-diagonal low-rank property, we examine one dataset,
GAS1K, whose kernel matrix $K$ has dimension $N=1000$. In
Figure~\ref{fig:svdecay}, we plot the singular values of the off-diagonal block $K(1,2)$ of size 500, with $h$ values varying from small to large. Figure~\ref{fig:sv-full} plots the singular values of the entire kernel matrix. We use both the natural ordering of the rows/columns
as well as a reordering of the rows/columns based on a recursive
two-means (2MN) clustering algorithm applied to the input data (see
Section~\ref{sec:clustering}).  For the same $K(1,2)$ block,
Table~\ref{tab:effective-sv} lists the number of singular values
larger than 0.01; we call this the effective rank.  As can be seen
from both Figure~\ref{fig:svdecay} and Table~\ref{tab:effective-sv},
The 2MN preprocessing leads to much faster decay for $h \sim 1$,
significantly improves the potential benefits of a solvers that
exploits the off-diagonal low-rank property. Reordering of the input
data is the main subject of this paper.	
	
\begin{table}[!h]
	\centering
	\begin{tabular}{c|ccccc}
	\hline
	h  & 0.01 & 0.1 & 1  & 10 & 100\\ \hline
	effective rank N/P & 1 & 23  & 338   &  129 & 14 \\ \hline
	effective rank 2MN & 1 & 1  & 78   &  76 & 12 \\ \hline
	\end{tabular}
	\caption{GAS1K: effective rank = number of singular values of the off-diagonal
	$500\times 500$ $K(1,2)$ block that are $> 0.01$, with and without
	2MN clustering. Effective rank gets small when $h \to 0$ or $h \to \infty$.}
	\label{tab:effective-sv}
\end{table}
    
\subsection{Main Contributions}

We propose to use hierarchical matrix approximations for the
kernel. We use both Hierarchically Semi-Separable (HSS) as well as
$\mathcal{H}$ matrices. The HSS algorithms are implemented as part of
the STRUMPACK library. STRUMPACK --- STRUctured Matrix PACKage --- is
a fast linear solver and preconditioner for sparse and dense systems.
One salient feature of STRUMPACK is that the HSS construction uses
adaptive randomized sampling for numerical rank detection, which
requires a black-box matrix times vector multiplication routine as
well as access to selected elements from the kernel
matrix~\eqref{K}. There is no need to explicitly store the whole
matrix $K$; we call this the {\em partially matrix-free} interface.
This feature is particularly attractive for kernel methods, because
forming the complete $K$ may consume too much memory.

Although STRUMPACK works algebraically with any input matrix,
different row and column orderings affect the numerical ranks of the
off-diagonal blocks, and hence the performance.  Intuitively, for an
off-diagonal block $K(I,J)$ to have a small rank, the data points in
the clusters $I$ and $J$ need to have as little interaction as
possible. This suggests a preprocessing step using clustering
algorithms, namely, (a) find groups of points with large inter-group
distances and small intra-group distances, and (b) permute the rows
and columns of the matrix $K$ such that the points of each group have
consecutive indices.
	
Another performance-critical aspect is how the random sampling is
performed. If we use traditional matrix-matrix multiplication to
perform the sampling $K R$, where $R$ consists of a number of random
vectors, the entire solution time will be dominated by this
operation. Here, we can exploit the special structure of matrix $K$
and use a faster structured sampling method with an $\mathcal{H}$
matrix.
	
To summarize, the main contributions of this work are:

\begin{itemize}
	\item We present an algorithm allowing fast approximate kernel matrix computations with linear scalability of the factorization and solution phases.
	\item For the preprocessing step, we explore various row and column orderings to improve the efficiency of HSS approximations in kernel methods.
	\item We use a fast sampling method based on the $\mathcal H$ matrix approximation of $K$, removing the bottleneck of the sampling phase for the HSS construction.
	\item We use the auto-tuning framework OpenTuner for the tuning of the hyperparameters in classification using kernel ridge regression.
	\item We report scalability experiments up to 1,024 cores, with real-world datasets from the UCI \cite{uci} up to $N$=4.5M data points for training, and dimension 784.
\end{itemize}
	
We experimented with a number of clustering techniques (including
versions of agglomerative and hierarchical clusterings, as well as
divisive 2-means, kd-tree and PCA-tree clusterings) and achieved
improvements up to $10 \times$ in terms of memory usage, with much
lower ranks, of the compressed matrix compared to the naive
application of STRUMPACK (without clustering) and up to $4 \times$
compared to k-d tree based reordering.  We believe that the clustering
techniques might be useful for more general classes of data related
matrices, in order to reveal the implicit hierarchical off-diagonal
low-rank structure, and will make them amenable to fast and scalable
algorithms. Asymptotically quasi-optimal memory consumption is key for the kernel ridge regression to be able to process large datasets.
	
\subsection{Previous Work}

A number of methods have been developed in the intersection of scientific
computing and machine learning, with the goal to accelerate kernel methods.
Here we give some highlights of the important methods, although we do not
have an exhaustive list covering the entire field.

The best rank $r$ approximation of a matrix would be given by direct
low-rank matrix factorization methods, based on singular value decomposition.
However, even with multiple improvements (in particular, due to the use of
randomization~\cite{Sarlos}, see also a survey \cite{Mah_survey}),
the total complexity of such methods stays quadratic in the number of
samples $\mathcal{O}(n^2\log r)$. They also require creating and storing the
entire kernel matrix $K$.

When the kernel matrix exhibits globally low rank, Nystr\"{o}m methods
are shown to be among the best methods (see, for example, \cite{GM, WS}).
Unfortunately, not all kernel matrices can be well approximated by
low-rank matrices in a global sense.

Other approximation approaches include fast multipole method (FMM)
partitions that try to split the data points into different boxes and
quantify interactions between the points that are far apart
(such as \cite{FMM1}) -- however, it seems that these methods work well
only for low dimensional data. 
	
The next idea, also crucial for our work, is to combine the low rank approximation with a clustering of the data points. Some of the previous algorithms using this idea are the following. Clustered Low-Rank Approximation (CLRA, \cite{SD} and its parallel version \cite{Sui}) was created to process large graphs. It starts with the clustering of the adjacency matrix, and then computes a low-rank approximation of each cluster (i.e., diagonal block), e.g. using singular value decomposition. Memory Efficient Kernel Representation (MEKA, \cite{Si})  also performs clustering, and then applies Nystr\"om approximation to within-cluster blocks (to avoid computing all within block entries), and introduces a sampling approach to capture between-block information. Block Basis Factorization (BFF, \cite{Mahoney}) improves upon MEKA in several ways, including computation of the low-rank basis vectors from a larger space, introducing a more sophisticated sampling procedure, and estimating near-optimal $k$, the number of clusters for the initial data splitting.

Our approach exploits both clustering and low rank property, but in a different way. Instead of initial splitting of data into $k$ clusters, we construct a binary tree of embedded clusters. Then we use low rank property of the off-diagonal sub-blocks of the kernel matrix, corresponding to the inter-cluster links, as well as the hierarchical connection between the clusters (and, respectively, sub-blocks) for efficient approximation of the low rank bases.
    
The line of work that is closest to ours and also based on off-diagonal
low rank is presented in a series of
papers~\cite{Biros,Biros-2,Biros-3,Biros-4,Biros-5},
where the block-diagonal-plus-low-rank hierarchical matrix format
is used to approximate the kernel matrix. The authors first developed an 
$\mathcal{O}(d N \log N)$ algorithm ASKIT to construct the approximate
representation for the kernel matrix~\cite{Biros-2}, and later
an $\mathcal{O}( N \log N)$ algorithm INV-ASKIT to perform a
factorization of the approximate matrix~\cite{Biros-3,Biros-5},
which can be used as a direct linear solver.
	
Our current work differs from the INV-ASKIT approach in several ways:

1) we use the $\mathcal{H}$ and HSS matrix formats in the approximation, 2) we use ULV
factorization, rather than the Sherman-Morrison-Woodbury
formula used in INV-ASKIT, and 3) we compare a number of
clustering methods to help reduce the off-diagonal rank, while
INV-ASKIT only used the k-d tree ordering.
	
\section{Kernel Ridge Regression for Classification}
    
Ridge regression is probably the most elementary algorithm that can be \textit{kernelized}. Classical ridge regression is designed to find the linear hyperplane that approximates the data labels well, and at the same time does not have too large coefficients, namely
$$
\argmin_{w} \sum_{i = 1}^n (y_i - w\cdot x_i)^2 + \lambda \|w\|_2^2,
$$
where $x_i$ are data points (rows of the data matrix $X$), $y_i$'s are their labels, $y = (y_1, \ldots, y_n)$ and $w$ is the normal vector to the target hyperplane. It can be proved (see, for example, \cite{hastie}) that the optimal $w$ is given by
$$
\hat w := X^T (X X^T + \lambda I)^{-1} y.
$$
The kernel trick introduces a way to replace the matrix $X X^T$ by a certain kernel matrix $K$, effectively substituting the scalar products $x_i\cdot x_j = X_{ij}$ by the elements $K(x_i, x_j) = K_{ij}$, that represent the scalar product in some higher dimensional space. 
    
Specifically, our experimental results for the two-class classification using ridge regression with the Gaussian kernel~(\ref{K}) are obtained by the following Algorithm~\ref{alg:ridge-regression}:

\begin{algorithm}
  \textbf{Input}: $X$ -- $n \times d$ train data matrix; \\
  \qquad\quad \ \  $X'$ -- $m \times d$ test data matrix; \\
  \qquad\quad \ \  $y \in \{\pm 1\}^n$ -- train labels \\
  \textbf{Output}: $y' \in \{\pm 1\}^m$ -- predicted test labels\\
  \textbf{Parameters:} $h$ -- Gaussian width; \\
  \qquad\qquad\qquad \textbf{$\lambda$} -- regularization parameter \\
  \textbf{0.} Preprocessing step: reorder data points $x_i$'s so that $dist(x_i, x_j)$ is small if and only if $|i - j|$ is small (see Section \ref{sec:clustering} for the details).\\
  \textbf{1.} Compute kernel matrix on the train data. \\
  $
  \quad K_{ij} := K(x_i, x_j) = \exp( -\|x_i - x_j\|^2/(2h^2))
  $ \\
  \quad where data points $x_i$ are the rows of $X$, $i = 1, \ldots, n$. \\
  \textbf{2.} Compute weight vector $w$ by solving the linear system \\
  $
  \quad  w := (K + \lambda I)^{-1}  y
  $   
  \\
  \textbf{3.} For each test data sample $x'_i \in X'$, $i = 1, \ldots, m$, \\
  compute kernel vector w.r.t. the train data \\
  $
  \quad K'(i)_j := \exp( -\|x_j - x'_i\|^2/(2h^2)), \quad j = 1, \ldots n,
  $ \\
  $
  \quad K'(i) := (K'(i)_1, \ldots, K'(i)_n)^T
  $
  \\
  \textbf{4.} For each $x'_i \in X'$, predict its class label as \\
  $
  \quad y'_i := \sign(w^T \cdot K'(i))
  $
\caption{\textbf{Gaussian kernel ridge regression}}
\label{alg:ridge-regression}
\end{algorithm}

The choice of parameters ($h$, $\lambda$) is based on a particular dataset and usually made by a cross-validation. The prediction accuracy is
computed as the fraction of correctly predicted test labels:
	\begin{equation}
	Acc = \frac{|\{y' \equiv \bar y'\}|}{m},
	\end{equation}
where $\bar y' \in \{\pm 1\}^m$ are the \emph{true} test labels (and $y' \in \{\pm 1\}^m$ are \emph{predicted} test labels).

In Algorithm~\ref{alg:ridge-regression}, the most time-consuming
computation is solving the linear system during the training stage
(Step 2).  One key observation is that the final prediction accuracy
does not require many digits of the solution weight vector $w$, since
it contributes only to the sign calculation in Step 4. Therefore, we
can use an inexact but faster linear solver, such as STRUMPACK, to
alleviate the performance bottleneck at Step 2. The goal of preprocessing Step 0 is to have  matrix $K$ (constructed on Step 1) in an HSS format, at minimum rank.
    
Other algorithms are also suggested for the kernel matrix compression
(see~\cite{GM,hastie}). In~\cite{GM} the authors show the
results of the binary classification by the kernel ridge regression.
There does not seem to be one best solution suggested so far.

A possible future work is to fully compare the classification accuracy
and the speed with STRUMPACK to that obtained from the method
described in~\cite{Biros-3}.
    
Finally, although Algorithm~\ref{alg:ridge-regression} is defined for two classes, it is easy to adapt it for the multi-class classification. The simplest way to do it is to make one-vs-all predictions. To distinguish between $c > 2$ classes, we would need to construct $c$ binary classifiers, that differ from the Algorithm~\ref{alg:ridge-regression} only in Step 4. Namely,
$$
 y'(c)_i := |w^T(c) \cdot K'(i)|,
$$ 
which interprets now as the level of confidence that $i$-th test point belongs to the class $c$. Then the class of $x'_i$ is defined as
$$c(i) := \argmax_c y'(c)_i.$$

\section{Data Sparse Formats}
In this section, we briefly describe two hierarchical matrix formats
used in this work, the hierarchically semi-separable (HSS)
representation, in Section~\ref{sec:HSS}, and the hierarchical
$\mathcal{H}$ matrix format, in Section~\ref{sec:H}. These matrix
representations use a hierarchical partitioning of the matrix into
smaller blocks, some of which can be compressed using low-rank
approximations.

The two hierarchical matrix formats play different roles in this work.
The HSS matrix has higher construction cost (if we use traditional
dense matrix multiplication in the sampling stage) but very low
factorization and solve cost, whereas the $\mathcal{H}$ matrix has
lower construction and matrix-vector multiplication cost but higher
factorization cost. Our approach is to use the fast matrix-vector
multiplication capability of the $\mathcal{H}$ matrix to speed up the
HSS matrix construction.
	
\subsection{Hierarchically Semi-Separable Matrix Representation\label{sec:HSS}}

\begin{figure}
		\centering
		\begin{minipage}{.47\columnwidth}
			\centering
			\begin{tikzpicture}[scale=3]
			\path[use as bounding box] (0,0) rectangle (1.2,1); 
			\node [label] at (0.25,0.25) {$A_{13,6}$};
			\node [label] at (0.75,0.75) {$A_{6,13}$};
			\draw (0,0) rectangle (1,1);
			\draw (0,0) rectangle (0.5,0.5);
			\draw (1,1) rectangle (0.5,0.5);
			\draw (0,1) rectangle (0.25,0.75);
			\draw [fill=gray] (0,1) rectangle (0.25/2,0.75+0.25/2);
			\draw [fill=gray] (0.25/2,0.75+0.25/2) rectangle (0.25,0.75);
			\draw (0.5,0.5) rectangle (0.75,0.25);
			\draw [fill=gray] (0.5,0.5) rectangle (0.75-0.25/2,0.25+0.25/2);
			\draw [fill=gray] (0.75-0.25/2,0.25+0.25/2) rectangle (0.75,0.25);
			\draw (0.25,0.75) rectangle (0.5,0.5);
			\draw [fill=gray] (0.25,0.75) rectangle (0.5-0.25/2,0.5+0.25/2);
			\draw [fill=gray] (0.5-0.25/2,0.5+0.25/2) rectangle (0.5,0.5);
			\draw (0.75,0.25) rectangle (1,0);
			\draw [fill=gray] (0.75,0.25) rectangle (0.75+0.25/2,0.25/2);
			\draw [fill=gray] (0.75+0.25/2,0.25/2) rectangle (1,0);
			\path [fill=gray] (0.02,0.35) rectangle (0.07,0.47);
			\path [fill=gray] (0.08,0.42) rectangle (0.13,0.47);
			\path [fill=gray] (0.14,0.42) rectangle (0.22,0.47);
			\path [fill=gray] (0.02+0.5,0.35+0.5) rectangle (0.07+0.5,0.47+0.5);
			\path [fill=gray] (0.08+0.5,0.42+0.5) rectangle (0.13+0.5,0.47+0.5);
			\path [fill=gray] (0.14+0.5,0.42+0.5) rectangle (0.22+0.5,0.47+0.5);
			\path [fill=gray] (0.25+0.01+0.5,0.01) rectangle (0.25+0.03+0.5,0.25/2-0.01);
			\path [fill=gray] (0.25+0.035+0.5,0.25/2-0.025-0.01) rectangle (0.25+0.055+0.5,0.25/2-0.01);
			\path [fill=gray] (0.25+0.06+0.5,0.25/2-0.025-0.01) rectangle (0.25+0.25/2-0.01+0.5,0.25/2-0.01);
			\path [fill=gray] (0.25/2+0.25+0.01+0.5,0.25/2+0.01) rectangle (0.25/2+0.25+0.03+0.5,0.25/2+0.25/2-0.01);
			\path [fill=gray] (0.25/2+0.25+0.035+0.5,0.25/2+0.25/2-0.025-0.01) rectangle (0.25/2+0.25+0.055+0.5,0.25/2+0.25/2-0.01);
			\path [fill=gray] (0.25/2+0.25+0.06+0.5,0.25/2+0.25/2-0.025-0.01) rectangle (0.25/2+0.25+0.25/2-0.01+0.5,0.25/2+0.25/2-0.01);
			\path [fill=gray] (0.01+0.5,0.01+0.25) rectangle (0.03+0.5,0.25/2-0.01+0.25);
			\path [fill=gray] (0.035+0.5,0.25/2-0.025-0.01+0.25) rectangle (0.055+0.5,0.25/2-0.01+0.25);
			\path [fill=gray] (0.06+0.5,0.25/2-0.025-0.01+0.25) rectangle (0.25/2-0.01+0.5,0.25/2-0.01+0.25);
			\path [fill=gray] (0.25/2+0.01+0.5,0.25/2+0.01+0.25) rectangle (0.25/2+0.03+0.5,0.25/2+0.25/2-0.01+0.25);
			\path [fill=gray] (0.25/2+0.035+0.5,0.25/2+0.25/2-0.025-0.01+0.25) rectangle (0.25/2+0.055+0.5,0.25/2+0.25/2-0.01+0.25);
			\path [fill=gray] (0.25/2+0.06+0.5,0.25/2+0.25/2-0.025-0.01+0.25) rectangle (0.25/2+0.25/2-0.01+0.5,0.25/2+0.25/2-0.01+0.25);
			\path [fill=gray] (-0.5+0.25+0.01+0.5,0.5+0.01) rectangle (-0.5+0.25+0.03+0.5,0.5+0.25/2-0.01);
			\path [fill=gray] (-0.5+0.25+0.035+0.5,0.5+0.25/2-0.025-0.01) rectangle (-0.5+0.25+0.055+0.5,0.5+0.25/2-0.01);
			\path [fill=gray] (-0.5+0.25+0.06+0.5,0.5+0.25/2-0.025-0.01) rectangle (-0.5+0.25+0.25/2-0.01+0.5,0.5+0.25/2-0.01);
			\path [fill=gray] (-0.5+0.25/2+0.25+0.01+0.5,0.5+0.25/2+0.01) rectangle (-0.5+0.25/2+0.25+0.03+0.5,0.5+0.25/2+0.25/2-0.01);
			\path [fill=gray] (-0.5+0.25/2+0.25+0.035+0.5,0.5+0.25/2+0.25/2-0.025-0.01) rectangle (-0.5+0.25/2+0.25+0.055+0.5,0.5+0.25/2+0.25/2-0.01);
			\path [fill=gray] (-0.5+0.25/2+0.25+0.06+0.5,0.5+0.25/2+0.25/2-0.025-0.01) rectangle (-0.5+0.25/2+0.25+0.25/2-0.01+0.5,0.5+0.25/2+0.25/2-0.01);
			\path [fill=gray] (-0.5+0.01+0.5,0.5+0.01+0.25) rectangle (-0.5+0.03+0.5,0.5+0.25/2-0.01+0.25);
			\path [fill=gray] (-0.5+0.035+0.5,0.5+0.25/2-0.025-0.01+0.25) rectangle (-0.5+0.055+0.5,0.5+0.25/2-0.01+0.25);
			\path [fill=gray] (-0.5+0.06+0.5,0.5+0.25/2-0.025-0.01+0.25) rectangle (-0.5+0.25/2-0.01+0.5,0.5+0.25/2-0.01+0.25);
			\path [fill=gray] (-0.5+0.25/2+0.01+0.5,0.5+0.25/2+0.01+0.25) rectangle (-0.5+0.25/2+0.03+0.5,0.5+0.25/2+0.25/2-0.01+0.25);
			\path [fill=gray] (-0.5+0.25/2+0.035+0.5,0.5+0.25/2+0.25/2-0.025-0.01+0.25) rectangle (-0.5+0.25/2+0.055+0.5,0.5+0.25/2+0.25/2-0.01+0.25);
			\path [fill=gray] (-0.5+0.25/2+0.06+0.5,0.5+0.25/2+0.25/2-0.025-0.01+0.25) rectangle (-0.5+0.25/2+0.25/2-0.01+0.5,0.5+0.25/2+0.25/2-0.01+0.25);
			\path [fill=gray] (0.015,0.15+0.5) rectangle (0.05,0.235+0.5);
			\path [fill=gray] (0.06,0.2+0.5) rectangle (0.09,0.235+0.5);
			\path [fill=gray] (0.1,0.2+0.5) rectangle (0.16,0.235+0.5);
			\path [fill=gray] (0.015+0.25,0.15+0.75) rectangle (0.05+0.25,0.235+0.5+0.25);
			\path [fill=gray] (0.06+0.25,0.2+0.75) rectangle (0.09+0.25,0.235+0.5+0.25);
			\path [fill=gray] (0.1+0.25,0.2+0.75) rectangle (0.16+0.25,0.235+0.5+0.25);
			\path [fill=gray] (0.5+0.015,0.15+0.5-0.5) rectangle (0.5+0.05,0.235+0.5-0.5);
			\path [fill=gray] (0.5+0.06,0.2+0.5-0.5) rectangle (0.5+0.09,0.235+0.5-0.5);
			\path [fill=gray] (0.5+0.1,0.2+0.5-0.5) rectangle (0.5+0.16,0.235+0.5-0.5);
			\path [fill=gray] (0.5+0.015+0.25,0.15+0.75-0.5) rectangle (0.5+0.05+0.25,0.235+0.5+0.25-0.5);
			\path [fill=gray] (0.5+0.06+0.25,0.2+0.75-0.5) rectangle (0.5+0.09+0.25,0.235+0.5+0.25-0.5);
			\path [fill=gray] (0.5+0.1+0.25,0.2+0.75-0.5) rectangle (0.5+0.16+0.25,0.235+0.5+0.25-0.5);
			\node [label] at (0.125/2*6,1-0.125/2*2) {\scriptsize $A_{2,5}$};
			\node [label] at (0.125/2*2,1-0.125/2*6) {\scriptsize $A_{5,2}$};
			\node [label] at (0.125/2*10,1-0.125/2*14) {\scriptsize $A_{12,9}$};
			\node [label] at (0.125/2*14,1-0.125/2*10) {\scriptsize $A_{9,12}$};
			\node [label] at (0.125/2,1-0.125/2) {\scriptsize $D_{0}$};
			\node [label] at (0.125/2*3,1-0.125/2*3) {\scriptsize $D_{1}$};
			\node [label] at (0.125/2*5,1-0.125/2*5) {\scriptsize $D_{3}$};
			\node [label] at (0.125/2*7,1-0.125/2*7) {\scriptsize $D_{4}$};
			\node [label] at (0.125/2*9,1-0.125/2*9) {\scriptsize $D_{7}$};
			\node [label] at (0.125/2*11,1-0.125/2*11) {\scriptsize $D_{8}$};
			\node [label] at (0.125/2*13,1-0.125/2*13) {\scriptsize $D_{10}$};
			\node [label] at (0.125/2*15,1-0.125/2*15) {\scriptsize $D_{11}$};
			\draw [decorate,decoration={brace,amplitude=2pt},yshift=0pt,xshift=-.1cm] (1.125,1-0.125*0) -- (1.125,1-0.125) node [black,midway,xshift=.3cm] {\footnotesize $I_0$};
			\draw [decorate,decoration={brace,amplitude=2pt},yshift=0pt,xshift=-.1cm] (1.125,1-0.125*1) -- (1.125,1-0.125*2) node [black,midway,xshift=.3cm] {\footnotesize $I_1$};
			\draw [decorate,decoration={brace,amplitude=2pt},yshift=0pt,xshift=-.1cm] (1.125,1-0.125*2) -- (1.125,1-0.125*3) node [black,midway,xshift=.3cm] {\footnotesize $I_3$};
			\draw [decorate,decoration={brace,amplitude=2pt},yshift=0pt,xshift=-.1cm] (1.125,1-0.125*3) -- (1.125,1-0.125*4) node [black,midway,xshift=.3cm] {\footnotesize $I_4$};
			\draw [decorate,decoration={brace,amplitude=2pt},yshift=0pt,xshift=-.1cm] (1.125,1-0.125*4) -- (1.125,1-0.125*5) node [black,midway,xshift=.3cm] {\footnotesize $I_7$};
			\draw [decorate,decoration={brace,amplitude=2pt},yshift=0pt,xshift=-.1cm] (1.125,1-0.125*5) -- (1.125,1-0.125*6) node [black,midway,xshift=.3cm] {\footnotesize $I_8$};
			\draw [decorate,decoration={brace,amplitude=2pt},yshift=0pt,xshift=-.1cm] (1.125,1-0.125*6) -- (1.125,1-0.125*7) node [black,midway,xshift=.3cm] {\footnotesize $I_{10}$};
			\draw [decorate,decoration={brace,amplitude=2pt},yshift=0pt,xshift=-.1cm] (1.125,1-0.125*7) -- (1.125,1-0.125*8) node [black,midway,xshift=.3cm] {\footnotesize $I_{11}$};
			\end{tikzpicture}
			\caption{\footnotesize Illustration of an HSS matrix using $4$
				levels. Diagonal blocks are partitioned recursively. Gray blocks
				denote the basis matrices.}
			\label{fig:HSS}
		\end{minipage} \hfill
		\begin{minipage}{.50\columnwidth}
			\centering
			\begin{tikzpicture}\scriptsize
			\node[cnode](14){14}
			child{node[cnode](6){6}
				child{node[cnode](2){2}   child{node[cnode](0){0}} child{node[cnode](1){1}} }
				child{node[cnode](5){5}   child{node[cnode](3){3}} child{node[cnode](4){4}} }
			}
			child{node[cnode](13){13}
				child{node[cnode](9){9}   child{node[cnode](7){7}}   child{node[cnode](8){8}} }
				child{node[cnode](12){12} child{node[cnode](10){10}} child{node[cnode](11){11}} }
			};
			\end{tikzpicture}
			\caption{\footnotesize Tree for Figure~\ref{fig:HSS}, using
				postordering. All nodes except the root store $U_i$ and
				$V_i$. Leaves store $D_i$, non-leaves $B_{ij}$, $B_{ji}$}
			\label{fig:HSStree}
		\end{minipage}
	\end{figure}
    
The HSS representation, as illustrated in Figure~\ref{fig:HSS}, uses a
block $2 \times 2$ partitioning of a matrix
$A \in \mathbb{R}^{n \times n}$, with a similar partitioning applied
recursively to the diagonal blocks. This recursive partitioning
defines a tree, as illustrated in Figure~\ref{fig:HSStree}. With a
node $i$ in the tree, an index set
$I_i \subset \left\{1,\dots,n\right\}$ is associated. At the last
level of the recursion, the diagonal blocks, i.e., $A(I_i, I_i)$ are
stored as (small) dense matrices. All off-diagonal blocks $A_{ij}$ are
compressed using a low-rank factorization $U_i B_{ij}
V_j^T$. Moreover, the column basis matrix $U_i$, for a node $i$ with
children $c_1$ and $c_2$ in the hierarchy is defined as
$U_i = \begin{bmatrix} U_{c_1} & 0 \\ 0 & U_{c_2} \end{bmatrix}
\tilde{U}_i$, and hence only the smaller matrix $\tilde{U}_i$ is
stored at node $i$. Only at the leaf nodes, where
$U_i \equiv \tilde{U}_i$, are the $U_i$ stored explicitly. A similar
relation holds for the $V_i$ basis matrices, and is referred to as the
nested basis property. The HSS data structure is implemented in the
STRUMPACK (STRUctured Matrix PACKage)
library~\cite{ghysels2017robust,1,3}. STRUMPACK is a sparse direct
solver and preconditioner. It uses HSS compression for the sparse
triangular factors. However, the HSS kernels implemented in STRUMPACK
can also be used directly on dense matrices, for instance coming from
integral equations, the boundary element method, electromagnetic
scattering etc. For the construction of HSS matrices, STRUMPACK
implements a randomized algorithm from~\cite{2}. This algorithm
requires a matrix times (multiple) vector product for the random
sampling phase. A fast multiplication routine is crucial to get good
performance. STRUMPACK also implements a ULV
factorization~\cite{chandrasekaran2006fast} algorithm, and a
corresponding routine to solve a linear system with the factored HSS
matrix. If a fast sampling routine is available, both HSS compression
and factorization have $\mathcal{O}(rn)$ complexity, with $r$ the
maximum HSS rank.
	
\subsection{$\mathcal{H}$ Matrix Representation \label{sec:H}}

        \begin{figure}[]
          \centering
          \begin{minipage}{.4\columnwidth}
            \centering
            \begin{tikzpicture}[scale=3]
              \path[use as bounding box] (0,0) rectangle (1,1); 
              \draw (0,0) rectangle (1,1);
              \draw (0,0) rectangle (0.5,0.5);
              \draw (1,1) rectangle (0.5,0.5);
              \draw (0,1) rectangle (0.25,0.75);
              \draw (0.5,0.5) rectangle (0.75,0.25);
              \draw (0.25,0.75) rectangle (0.5,0.5);
              \draw (0.75,0.25) rectangle (1,0);
              \draw (0.5,0.5) rectangle (0.75,0.75);
              \draw (0.75,0.75) rectangle (1,1);

              \draw [fill=gray] (0.5+0.25/2,0.25/2) rectangle (0.75,0.25);
              \draw (0.5,0) rectangle (0.5+0.25/2,0.25/2);

              \draw [fill=gray] (0.5,0.5) rectangle (0.25+0.25/2+0.25,0.25+0.25/2+0.25);
              \draw [fill=gray] (0.5+0.25,0.5-0.25) rectangle (0.25+0.25/2+0.5,0.25+0.25/2);

              \draw [fill=gray] (0,1) rectangle (0.25/2,1-0.25/2);
              \draw [fill=gray] (0.25/2,1-0.25/2) rectangle (0.25,0.75);
              \draw [fill=gray] (0.25,0.75) rectangle (0.25+0.25/2,0.75-0.25/2);
              \draw [fill=gray] (0.25+0.25/2,0.75-0.25/2) rectangle (0.5,0.5);
              \draw [fill=gray] (0.5,0.5) rectangle (0.5+0.25/2,0.5-0.25/2);
              \draw [fill=gray] (0.5+0.25/2,0.5-0.25/2) rectangle (0.75,0.25);
              \draw [fill=gray] (0.75,0.25) rectangle (0.75+0.25/2,0.25/2);
              \draw [fill=gray] (0.75+0.25/2,0.25/2) rectangle (1,0);

              \draw (0.75,0.75) rectangle (0.75-0.25/2,0.75-0.25/2);
              \draw (1,0.5) rectangle (1-0.25/2,0.5-0.25/2);

              \draw [fill=gray] (1,1) rectangle (0.75+0.25/2,0.75+0.25/2);
              \draw (0.75,0.75) rectangle (0.75+0.25/2,0.75+0.25/2);

              \path [fill=gray] (0.02,0.02+0.25) rectangle (0.05,0.225+0.25);
              \path [fill=gray] (0.07,0.195+0.25) rectangle (0.225,0.225+0.25);
              \path [fill=gray] (0.02+0.25,0.02) rectangle (0.05+0.25,0.225);
              \path [fill=gray] (0.07+0.25,0.195) rectangle (0.225+0.25,0.225);

              \path [fill=gray] (0.02,0.02+.5-0.25/2-0.25) rectangle (0.04,0.25/2-0.02+.5-0.25/2-0.25);
              \path [fill=gray] (0.05,0.25/2-0.02-0.02+.5-0.25/2-0.25) rectangle (0.25/2-0.02,0.25/2-0.02+.5-0.25/2-0.25);
              \path [fill=gray] (0.02+0.25/2,0.02+.5-0.25/2-0.25) rectangle (0.04+0.25/2,0.25/2-0.02+.5-0.25/2-0.25);
              \path [fill=gray] (0.05+0.25/2,0.25/2-0.02-0.02+.5-0.25/2-0.25) rectangle (0.25/2-0.02+0.25/2,0.25/2-0.02+.5-0.25/2-0.25);
              \path [fill=gray] (0.02+0.25/2,0.02+.5-0.25/2-0.25/2-0.25) rectangle (0.04+0.25/2,0.25/2-0.02+.5-0.25/2-0.25/2-0.25);
              \path [fill=gray] (0.05+0.25/2,0.25/2-0.02-0.02+.5-0.25/2-0.25/2-0.25) rectangle (0.25/2-0.02+0.25/2,0.25/2-0.02+.5-0.25/2-0.25/2-0.25);

              \path [fill=gray] (0.02+0.25,0.02+0.25) rectangle (0.25+0.04,0.25/2-0.02+0.25);
              \path [fill=gray] (0.05+0.25,0.25/2-0.02-0.02+0.25) rectangle (0.25+0.25/2-0.02,0.25/2-0.02+0.25);
              \path [fill=gray] (0.02+0.25+0.25/2,0.02+0.25) rectangle (0.25+0.04+0.25/2,0.25/2-0.02+0.25);
              \path [fill=gray] (0.05+0.25+0.25/2,0.25/2-0.02-0.02+0.25) rectangle (0.25+0.25/2-0.02+0.25/2,0.25/2-0.02+0.25);
              \path [fill=gray] (0.02+0.25,0.02+0.25+0.25/2) rectangle (0.25+0.04,0.25/2-0.02+0.25+0.25/2);
              \path [fill=gray] (0.05+0.25,0.25/2-0.02-0.02+0.25+0.25/2) rectangle (0.25+0.25/2-0.02,0.25/2-0.02+0.25+0.25/2);

              \draw (0.25,0.25) rectangle (0.25-0.25/2,0.25-0.25/2);
              \draw (0.25,0.25) rectangle (0,0);
              \draw [fill=gray] (0,0) rectangle (0.25/2,0.25/2);

              \draw (0.25,0.25) rectangle (0.25+0.25/2,0.25+0.25/2);
              \draw (0.25,0.25) rectangle (0.5,0.5);
              \draw [fill=gray] (0.25+0.25/2,0.25+0.25/2) rectangle (0.5,0.5);

              \path [fill=gray] (0.02,0.02+0.5) rectangle (0.05,0.225+0.5);
              \path [fill=gray] (0.07,0.195+0.5) rectangle (0.225,0.225+0.5);
              \path [fill=gray] (0.02+0.25,0.02+0.75) rectangle (0.05+0.25,0.225+0.75);
              \path [fill=gray] (0.07+0.25,0.195+0.75) rectangle (0.225+0.25,0.225+0.75);
              \path [fill=gray] (0.02+0.5,0.02+0.75) rectangle (0.05+0.5,0.225+0.75);
              \path [fill=gray] (0.07+0.5,0.195+0.75) rectangle (0.225+0.5,0.225+0.75);
              \path [fill=gray] (0.02+0.75,0.02+0.5) rectangle (0.05+0.75,0.225+0.5);
              \path [fill=gray] (0.07+0.75,0.195+0.5) rectangle (0.225+0.75,0.225+0.5);

              \path [fill=gray] (0.02+0.75,0.02+0) rectangle (0.04+0.75,0.25/2-0.02+0);
              \path [fill=gray] (0.05+0.75,0.25/2-0.02-0.02+0) rectangle (0.25/2-0.02+0.75,0.25/2-0.02+0);
              \path [fill=gray] (0.02+0.75+0.25/2,0.02+0+0.25/2) rectangle (0.04+0.75+0.25/2,0.25/2-0.02+0+0.25/2);
              \path [fill=gray] (0.05+0.75+0.25/2,0.25/2-0.02-0.02+0+0.25/2) rectangle (0.25/2-0.02+0.75+0.25/2,0.25/2-0.02+0+0.25/2);
              \path [fill=gray] (0.02+0.5,0.02+0+0.25) rectangle (0.04+0.5,0.25/2-0.02+0+0.25);
              \path [fill=gray] (0.05+0.5,0.25/2-0.02-0.02+0+0.25) rectangle (0.25/2-0.02+0.5,0.25/2-0.02+0+0.25);
              \path [fill=gray] (0.02+0.5+0.25/2,0.02+0+0.25+0.25/2) rectangle (0.04+0.5+0.25/2,0.25/2-0.02+0+0.25+0.25/2);
              \path [fill=gray] (0.05+0.5+0.25/2,0.25/2-0.02-0.02+0+0.25+0.25/2) rectangle (0.25/2-0.02+0.5+0.25/2,0.25/2-0.02+0+0.25+0.25/2);

              \path [fill=gray] (0.02,0.02+0.75) rectangle (0.04,0.25/2-0.02+0.75);
              \path [fill=gray] (0.05,0.25/2-0.02-0.02+0.75) rectangle (0.25/2-0.02,0.25/2-0.02+0.75);
              \path [fill=gray] (0.02+0.25/2,0.02+0.75+0.25/2) rectangle (0.04+0.25/2,0.25/2-0.02+0.75+0.25/2);
              \path [fill=gray] (0.05+0.25/2,0.25/2-0.02-0.02+0.75+0.25/2) rectangle (0.25/2-0.02+0.25/2,0.25/2-0.02+0.75+0.25/2);
              \path [fill=gray] (0.02+0.25,0.02+0.5) rectangle (0.04+0.25,0.25/2-0.02+0.5);
              \path [fill=gray] (0.05+0.25,0.25/2-0.02-0.02+0.5) rectangle (0.25/2-0.02+0.25,0.25/2-0.02+0.5);
              \path [fill=gray] (0.02+0.25+0.25/2,0.02+0.5+0.25/2) rectangle (0.04+0.25+0.25/2,0.25/2-0.02+0.5+0.25/2);
              \path [fill=gray] (0.05+0.25+0.25/2,0.25/2-0.02-0.02+0.5+0.25/2) rectangle (0.25/2-0.02+0.25+0.25/2,0.25/2-0.02+0.5+0.25/2);

              \path [fill=gray] (0.02+0.75,0.02+0.75) rectangle (0.75+0.04,0.25/2-0.02+0.75);
              \path [fill=gray] (0.05+0.75,0.25/2-0.02-0.02+0.75) rectangle (0.75+0.25/2-0.02,0.25/2-0.02+0.75);
              \path [fill=gray] (0.02+0.75+0.25/2,0.02+0.75) rectangle (0.75+0.04+0.25/2,0.25/2-0.02+0.75);
              \path [fill=gray] (0.05+0.75+0.25/2,0.25/2-0.02-0.02+0.75) rectangle (0.75+0.25/2-0.02+0.25/2,0.25/2-0.02+0.75);
              \path [fill=gray] (0.02+0.75,0.02+0.75+0.25/2) rectangle (0.75+0.04,0.25/2-0.02+0.75+0.25/2);
              \path [fill=gray] (0.05+0.75,0.25/2-0.02-0.02+0.75+0.25/2) rectangle (0.75+0.25/2-0.02,0.25/2-0.02+0.75+0.25/2);

              \path [fill=gray] (0.02+0.5,0.02+0) rectangle (0.04+0.5,0.25/2-0.02+0);
              \path [fill=gray] (0.05+0.5,0.25/2-0.02-0.02+0) rectangle (0.25/2-0.02+0.5,0.25/2-0.02+0);
              \path [fill=gray] (0.02+0.5+0.25/2,0.02+0) rectangle (0.04+0.5+0.25/2,0.25/2-0.02+0);
              \path [fill=gray] (0.05+0.5+0.25/2,0.25/2-0.02-0.02+0) rectangle (0.25/2-0.02+0.5+0.25/2,0.25/2-0.02+0);
              \path [fill=gray] (0.02+0.5,0.02+0+0.25/2) rectangle (0.04+0.5,0.25/2-0.02+0+0.25/2);
              \path [fill=gray] (0.05+0.5,0.25/2-0.02-0.02+0+0.25/2) rectangle (0.25/2-0.02+0.5,0.25/2-0.02+0+0.25/2);

              \path [fill=gray] (0.25+0.02+0.25,0.02+1-0.25/2-0.25) rectangle (0.25+0.04+0.25,0.25/2-0.02+1-0.25/2-0.25);
              \path [fill=gray] (0.25+0.05+0.25,0.25/2-0.02-0.02+1-0.25/2-0.25) rectangle (0.25+0.25/2-0.02+0.25,0.25/2-0.02+1-0.25/2-0.25);
              \path [fill=gray] (0.25+0.02+0.25+0.25/2,0.02+1-0.25/2-0.25) rectangle (0.25+0.04+0.25+0.25/2,0.25/2-0.02+1-0.25/2-0.25);
              \path [fill=gray] (0.25+0.05+0.25+0.25/2,0.25/2-0.02-0.02+1-0.25/2-0.25) rectangle (0.25+0.25/2-0.02+0.25+0.25/2,0.25/2-0.02+1-0.25/2-0.25);
              \path [fill=gray] (0.25+0.02+0.25+0.25/2,0.02+1-0.25/2-0.25/2-0.25) rectangle (0.25+0.04+0.25+0.25/2,0.25/2-0.02+1-0.25/2-0.25/2-0.25);
              \path [fill=gray] (0.25+0.05+0.25+0.25/2,0.25/2-0.02-0.02+1-0.25/2-0.25/2-0.25) rectangle (0.25+0.25/2-0.02+0.25+0.25/2,0.25/2-0.02+1-0.25/2-0.25/2-0.25);

              \path [fill=gray] (0.5+0.02+0.25,0.02+1-0.25/2-0.5) rectangle (0.5+0.04+0.25,0.25/2-0.02+1-0.25/2-0.5);
              \path [fill=gray] (0.5+0.05+0.25,0.25/2-0.02-0.02+1-0.25/2-0.5) rectangle (0.5+0.25/2-0.02+0.25,0.25/2-0.02+1-0.25/2-0.5);
              \path [fill=gray] (0.5+0.02+0.25+0.25/2,0.02+1-0.25/2-0.5) rectangle (0.5+0.04+0.25+0.25/2,0.25/2-0.02+1-0.25/2-0.5);
              \path [fill=gray] (0.5+0.05+0.25+0.25/2,0.25/2-0.02-0.02+1-0.25/2-0.5) rectangle (0.5+0.25/2-0.02+0.25+0.25/2,0.25/2-0.02+1-0.25/2-0.5);
              \path [fill=gray] (0.5+0.02+0.25+0.25/2,0.02+1-0.25/2-0.25/2-0.5) rectangle (0.5+0.04+0.25+0.25/2,0.25/2-0.02+1-0.25/2-0.25/2-0.5);
              \path [fill=gray] (0.5+0.05+0.25+0.25/2,0.25/2-0.02-0.02+1-0.25/2-0.25/2-0.5) rectangle (0.5+0.25/2-0.02+0.25+0.25/2,0.25/2-0.02+1-0.25/2-0.25/2-0.5);
            \end{tikzpicture}
          \end{minipage}
          \caption{Example of a symmetric $\mathcal{H}$ matrix using
            $4$ levels. Off-diagonal gray blocks denote low-rank
            factorizations.}
          \label{fig:h_image}
        \end{figure}
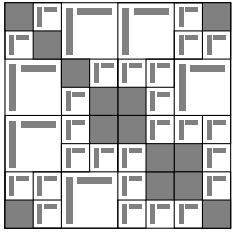

$\mathcal{H}$ (and $\mathcal{H}^2$) matrices are yet another group of
hierarchical matrix formats for building fast linear solvers. Contrary to
HSS, where all off-diagonal blocks are low-rank compressed
(weak admissibility), $\mathcal{H}$ formats only compress
the well-separated sub-blocks (strong admissibility). For example, the off-diagonal blocks in Figure~\ref{fig:h_image} are recursively partitioned into smaller ones and low-rank compressed if they are well-separated (i.e., admissible). Strong admissibility
leads to well-bounded numerical ranks for admissible blocks in linear
systems arising from high-dimensional applications (e.g., 3D acoustic 
and electromagnetic scattering problems and high-dimensional kernel matrices).
Therefore, the construction of an $\mathcal{H}$ matrix for the 
kernel matrix can be performed in quasi-linear time and memory.
However, it is well-known that the inversion of $\mathcal{H}$ matrices
requires significantly higher computation overhead when compared to
weak-admissibility solvers, such as those based on HSS.
Experiments showed that using an $\mathcal{H}$ solver to solve the
linear system in Step 2 of Algorithm~\ref{alg:ridge-regression} is much
slower than HSS due to the inversion bottleneck.
Therefore, instead of using the $\mathcal{H}$ solver to solve the system
directly, we use it only to compress the kernel matrix and then 
to accelerate the HSS construction as described below.

Recall that during the HSS construction, STRUMPACK requires rapid multiplication of the kernel matrix and its transpose to random vectors in the sampling stage.
Instead of direct application of the matrices to vectors, one can leverage Fast Gauss Transform \cite{FMM1}, analytical/algebraic fast multipole methods to accelerate the construction.
To this end, we tailored and adapted an $\mathcal{H}$ solver~\cite{H} to
kernel matrices. The low-rank representation of an admissible block is computed via a hybrid-ACA scheme that constructs a low-rank factorization of its submatrix that represents interactions between closely located points. The selection of the submatrix, however, is based on a trade-off between efficiency and
accuracy of the factorization. 	
	
\vspace{-10pt}
\section{Dataset Clustering}\label{sec:clustering}

This section discusses how appropriate preprocessing of the
input data can drastically improve the effectiveness of HSS
approximation. Section~\ref{sec:clusteringalgos} describes a
number of clustering algorithms to reduce the ranks of the HSS
off-diagonal blocks.

\subsection{Kernel Matrices and HSS Structure}

Intuitively, a kernel matrix can be viewed as a similarity matrix
$K = (K_{ij})_{i,j = 1, \ldots, n}$, where
\begin{equation}
K_{ij} = \text{ similarity score between } x_i \text{ and } x_j.
\end{equation}
Such a matrix $K$, defined by a set of objects $x_1, \ldots, x_n$ is
always square and symmetric. These objects can be anything from two
integers, two real valued vectors, to particles, words etc., provided
that we know how to compare them. Kernel matrices are used to improve
algorithms to classify these objects, 
e.g. allowing more sophisticated boundaries between classes. The Gaussian
kernel~\eqref{K}, studied throughout this paper, is but one example
kernel matrix; though probably the most widely used one. Let
$x_1, \ldots, x_n$ be data points in $d$ dimensional space $\R^d$,
which are said to be ``similar'' if they are close to each other in
Euclidean distance.

Broadly speaking, the preprocessing takes advantage of the fact that
the interaction between two well separated clusters of data points can
be approximated accurately when expressed in terms of the interaction
between a smaller number of representative points from each
cluster. This same idea is used very successfully in tree-codes \cite{Tree}, the
fast multipole method (FMM)~\cite{FMM}, matrix skeletonization \cite{Skel} and
interpolative decomposition \cite{ID}. Hence, splitting the input data in
clusters with large inter cluster distances leads to lower ranks for
the off-diagonal blocks of $K$, and hence less memory usage and faster
algorithms. This low rank property is illustrated in
Figure~\ref{fig:svdecay} with and without preprocessing, and it
motivates the idea of using an HSS solver like STRUMPACK to speed up
kernel matrix computations.  Due to the exponential decay of the
Gaussian kernel, many elements in the kernel matrix, away from the
diagonal, are actually negligibly small. Furthermore, the low rank pattern is
proven in~\cite{RBF}.

\subsection{Preprocessing by Data Clustering}

Reordering the input data $x_1, \ldots, x_n$ corresponds to
applying a permutation symmetrically to the rows and columns
of the kernel matrix. Preprocessing is applied in the
following steps:

\begin{enumerate}
\item Partition the data points in two clusters with large inter-group distance and small intra-group distance.
	\item Reorder the input data such that points in the same group occupy consecutive indices, increasing the data-sparsity of the off-diagonal blocks in the corresponding kernel matrix.
	\item Use these two index ranges to partition the corresponding node in the HSS tree, as illustrated in Figure~\ref{fig:HSStree}. Repeat Steps 1 and 2 recursively for both partitions computed in Step 1.
\end{enumerate}

Step 1 suggests that this preprocessing is a \emph{clustering
problem}. The task of finding clusters of points is widely
studied, with numerous clustering algorithms described in the
literature, without there being one absolute best solution.
Different algorithms perform better for different applications
and clustering quality can often be traded for execution time
or memory usage (see, for example,~\cite{hastie}).

Our specific requirements for an optimal clustering come from
(a) the properties of the HSS data structure and (b) the general
goal to minimize memory usage of the hierarchical matrix data
structures. The latter requires a clustering algorithm that
does not construct the kernel matrix $K$, or an equivalent
$n \times n$ distance matrix, explicitly. It should also be
fast, preferably $\mathcal{O}(n)$ or $\mathcal{O}(n \log n)$.
The HSS partitioning is defined by a hierarchical structure; a
possibly unbalanced and incomplete tree. However, in order to
be able to exploit enough parallelism and to reduce memory
usage, the tree should be deep enough, i.e., have a small
enough maximum leaf size.

Rather than the standard dissimilarity metrics measuring
clustering quality, the following performance metrics are used
in this work:

\begin{itemize}
    \item \textbf{Memory (MB)}: the sum of the memory used by all the individual smaller matrices in the HSS structure: $D_i$, $U_i$, $V_i$, $B_{ij}$, $B_{ji}$ (see
    Section~\ref{sec:HSS}).
    \item \textbf{Accuracy of classification (\%)}: the percentage of correctly predicted labels in the test set (with the parameters $h$ and $\lambda$ chosen based on the validation set).
    \item \textbf{Time (s)}: the time required for compression into HSS form, for factorization of the HSS matrix, and for solution of the linear system.
    \item \textbf{Maximum rank}: the largest rank encountered in any of the off-diagonal blocks of the HSS structure.
\end{itemize}

\subsection{Selected Preprocessing
Methods\label{sec:clusteringalgos}}

Previous work on approximation of kernel matrices used
reorderings based on ball tree clustering, see for
instance~\cite{Biros,Mahoney}, or based on k-d tree
clustering, see~\cite{kdtree}.

We have compared a variety of different clustering techniques
and their variations, both divisive and agglomerative, on
several real world datasets. Each of the clustering algorithms
used in the numerical results section are divisive, i.e., they
use a top-down, recursive split of the input points into two
separated clusters. The recursive splitting continues while
clusters are bigger than a certain leaf size, chosen to be
$16$ for HSS. This leaf size is the size of the diagonal
blocks in the HSS structure. The leaf size should not affect
the accuracy of the hierarchical matrix representation, but it
affects the memory usage.

Agglomerative methods, in contrast to the divisive strategy, although very good at reducing memory and ranks of the HSS structure, did not show competitive performance.
We experimented with a variety of hierarchical clustering methods, and typical disadvantages arising were either very unbalanced class sizes, or lack of parallelism ($\mathcal{O}(n^2)$ scaling, requiring to construct and store the complete distance matrix).
        
For the experiments we consider four orderings:

\noindent
\textbf{No preprocessing (NP)}: This is the baseline to
compare with: the input is not reordered, no information about
mutual distances is used to permute the matrix. The HSS tree
is a complete binary tree, constructed by recursively
splitting index sets in two equal ($\pm 1$) parts.

\noindent
\textbf{Recursive two-means (2MN)} This special case of the
well-known $k$-means clustering algorithm is applied
recursively to define the HSS tree. $K$-means is an iterative
algorithm which works as follows. Pick two points (at random)
to represent two clusters; for each point in the data set,
compute the distance to those two points and find which one is
closer, assign points to the closest cluster; compute the
center off each cluster and take that as the new representative
point of the cluster; repeat until no points change cluster.
Typically only a few iterations are required. However, the
procedure is relatively sensitive to the choice of initial
cluster representatives. Initially, we pick one point randomly 
and select the second one with a probability proportional to 
the distance from the first one.

\noindent
\textbf{K-d tree (KD)} The data is split along the coordinate
dimension of maximum spread, at the mean value for that
coordinate. Splitting at the mean is sensitive to outliers,
and can lead to very unbalanced trees. Alternatively,
splitting at the median value always results in a well
balanced tree. However, using the mean leads to lower memory
usage and when the input data is normalized, the sensitivity
with respect to outliers is less pronounced. If the resulting
clusters are still too unbalanced, i.e., when
($100\cdot\text{size(cluster1)} < \text{size(cluster2)}$), we
fall back to splitting at the median. This clustering is
applied recursively, where at each step of the recursion, a
new direction of maximum spread is determined.

\noindent
\textbf{Principal component analysis (PCA)} At each step of
the recursive clustering, the data is split according to the
mean value in the projection onto the first principal
component (i.e. direction of the maximum spread). We expect
this to be a better clustering than the simpler k-d tree
method, at a somewhat higher cost.

\section{Numerical Results\label{sec:results}}  

Experiments were performed at NERSC's Cori supercomputer. Each Cori node has two sockets, each socket is a 16-core Intel Xeon Processor E5-2698 v3 (``Haswell'') processor at 2.3 GHz and 128 GB DDR4 memory.

\subsection{Datasets Description}
	
We use real-world datasets coming from life sciences, physical sciences and artificial intelligence. The reported datasets are: SUSY, HEPMASS (high-energy physics, Monte Carlo simulated kinematic properties of the particles in the accelerator), COVTYPE (predicting forest cover type from cartographic variables), GAS (measurements from chemical sensors to distinguish between gases with  different concentration levels), PEN and LETTER (handwritten digits and letters recognition). All mentioned datasets are taken from the UCI repository \cite{uci}. Finally, to illustrate the performance on a dataset with large dimension, we used the MNIST dataset of handwritten digits (extended 8M dataset, including shifts and rotations of the classical MNIST digits dataset) \cite{libsvm}.
	
For the datasets having multiple classes we perform one-vs-all prediction. In particular, in MNIST and PEN we predict digit 5, in LETTER we predict letter A, in COVTYPE -- type 3 (Ponderosa Pine), in GAS -- gas number 5. Prediction accuracy might differ significantly if one would predict some other class.
	
\subsection{Preprocessing Comparison}

        \begin{table}[tbp]
        \small
		\centering
		{\footnotesize
			\begin{tabular}{|c|c|c|c|c|c|c|}
				\hline
				\multirow{2}{*}{\textbf{Dataset (dim)}} 
				& \multicolumn{4}{c|}{\textbf{Memory (MB)}} & \multirow{2}{*}{\textbf{Acc}} \\\cline{2-5}
				& \multicolumn{1}{c|}{\textbf{N/P}} & \multicolumn{1}{c|}{\textbf{KD}} & \multicolumn{1}{c|}{\textbf{PCA}} & \multicolumn{1}{c|}{\textbf{\begin{tabular}[c]{@{}c@{}}2MN\end{tabular}}}& \\ \hline
				
				\textbf{SUSY (8)}  & \multirow{2}{*}{499}  & \multirow{2}{*}{344} & \multirow{2}{*}{242} &\multirow{2}{*}{190} & \multirow{2}{*}{80.1\%}  \\
				h = 1, \textbf{$\lambda$} = 4 &  &&&& \\ \hline
                				\textbf{LETTER (16)}  & \multirow{2}{*}{315} & \multirow{2}{*}{237} & \multirow{2}{*}{91} & \multirow{2}{*}{51} & \multirow{2}{*}{100\%} \\ 
				h = .5, \textbf{$\lambda$} = 1  &  &  &  &  & \\ \hline
				\textbf{PEN (16)}  & \multirow{2}{*}{445} & \multirow{2}{*}{227} & \multirow{2}{*}{133} & \multirow{2}{*}{58} & \multirow{2}{*}{99.8\%}  \\ 
				h = 1, \textbf{$\lambda$} = 1 &   & &  &  &  \\ \hline
                				\textbf{HEPMASS (27)}  & \multirow{2}{*}{577} & \multirow{2}{*}{505} & \multirow{2}{*}{542} & \multirow{2}{*}{435} & \multirow{2}{*}{91.1\%}  \\ 
				h = 1.5, \textbf{$\lambda$} = 2 &   &  &  &  &  \\ \hline
				\textbf{COVTYPE (54)}  & \multirow{2}{*}{655} & \multirow{2}{*}{344} & \multirow{2}{*}{120}  & \multirow{2}{*}{45} & \multirow{2}{*}{97.1\%} \\ 
				h = 1, \textbf{$\lambda$} = 1    &  & & &  &  \\ \hline
                				\textbf{GAS (128)}  & \multirow{2}{*}{264} & \multirow{2}{*}{65} & \multirow{2}{*}{29} & \multirow{2}{*}{25} & \multirow{2}{*}{99.5\%} \\ 
				h = 1.5, \textbf{$\lambda$} = 4  & &  &&  & \\ \hline
				\textbf{MNIST (784)}  & \multirow{2}{*}{40} & \multirow{2}{*}{164} & \multirow{2}{*}{43} & \multirow{2}{*}{36} & \multirow{2}{*}{97.2\%}  \\ 
				h = 4, \textbf{$\lambda$} = 3 &   &&  &  &  \\ \hline
			\end{tabular}
		}
		\caption{Memory usage is in MB, accuracy is in \% of the test size (1K for all datasets). All train sets have 10K samples, normalized to zero mean and standard deviation one.}
		\label{tab:clustering}
	\end{table}

\begin{figure}[]
    \centering\includegraphics[width=1.4in]{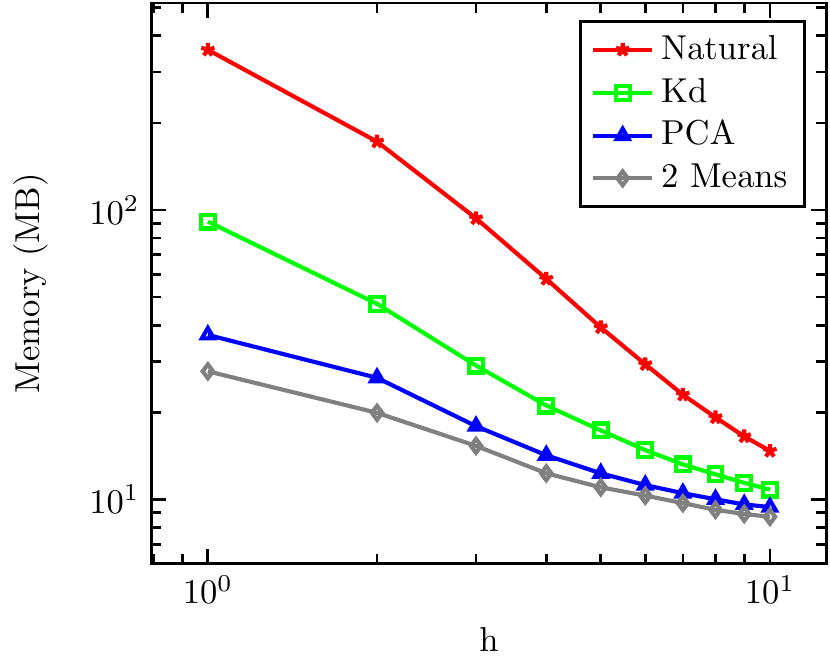}
    \caption{GAS10K dataset, 1K test, $\lambda = 4$. Memory usage for various $h$ and clustering methods.}
    \label{fig:GAS-h-memory}
\end{figure}

We report in Table~\ref{tab:clustering} our main performance metrics -- memory and accuracy. The table compares different preprocessing methods with seven datasets. Memory usage heavily depends on parameter $h$, which is illustrated on the GAS10K dataset in Fig.~\ref{fig:GAS-h-memory}. Furthermore, in the context of hierarchical low-rank approximations, memory is proportional to performance, since the number of flops is proportional to the numerical ranks of the approximation.
	
Recursive two means (2MN) preprocessing shows best memory performance 
for all the $h$ values.
With STRUMPACK tolerance set to be at most 0.1, the prediction accuracy
does not seem to depend on the preprocessing methods,
see Table~\ref{tab:clustering}.
Moreover, for the 10K datasets reported in Table~\ref{tab:clustering} this accuracy matches the accuracy we get using the full non-compressed kernel matrix in Algorithm~\ref{alg:ridge-regression}. The main disadvantage of 2MN is higher variance of the resulting rank, and, to lower extent, memory. The numbers reported for 2MN are average over three runs of the algorithm. The instability can be avoided by choosing non-random start points at every step of two means clustering. All datasets were normalized to have zero mean and unit standard deviation columns.	The experiments with non-normalized datasets, and with datasets normalized to have maximum absolute value one have shown significantly lower accuracy (e.g. for MNIST2M dataset).
	
\vspace{-10px}
\subsection{Hyperparameter Tuning}
    
The algorithm parameters $h$ and $\lambda$ are key to determine the
predictive capabilities of the matrix approximation. Consider the HSS
approximation of matrix $K + \lambda I$ used in this work.  When
the parameter $\lambda$ changes, we only need to update the diagonal entries
of the HSS matrix, and there is no need to perform HSS construction again.

\begin{figure}[]
	\begin{subfigure}{.23\textwidth}
		\includegraphics[height=1.5in]{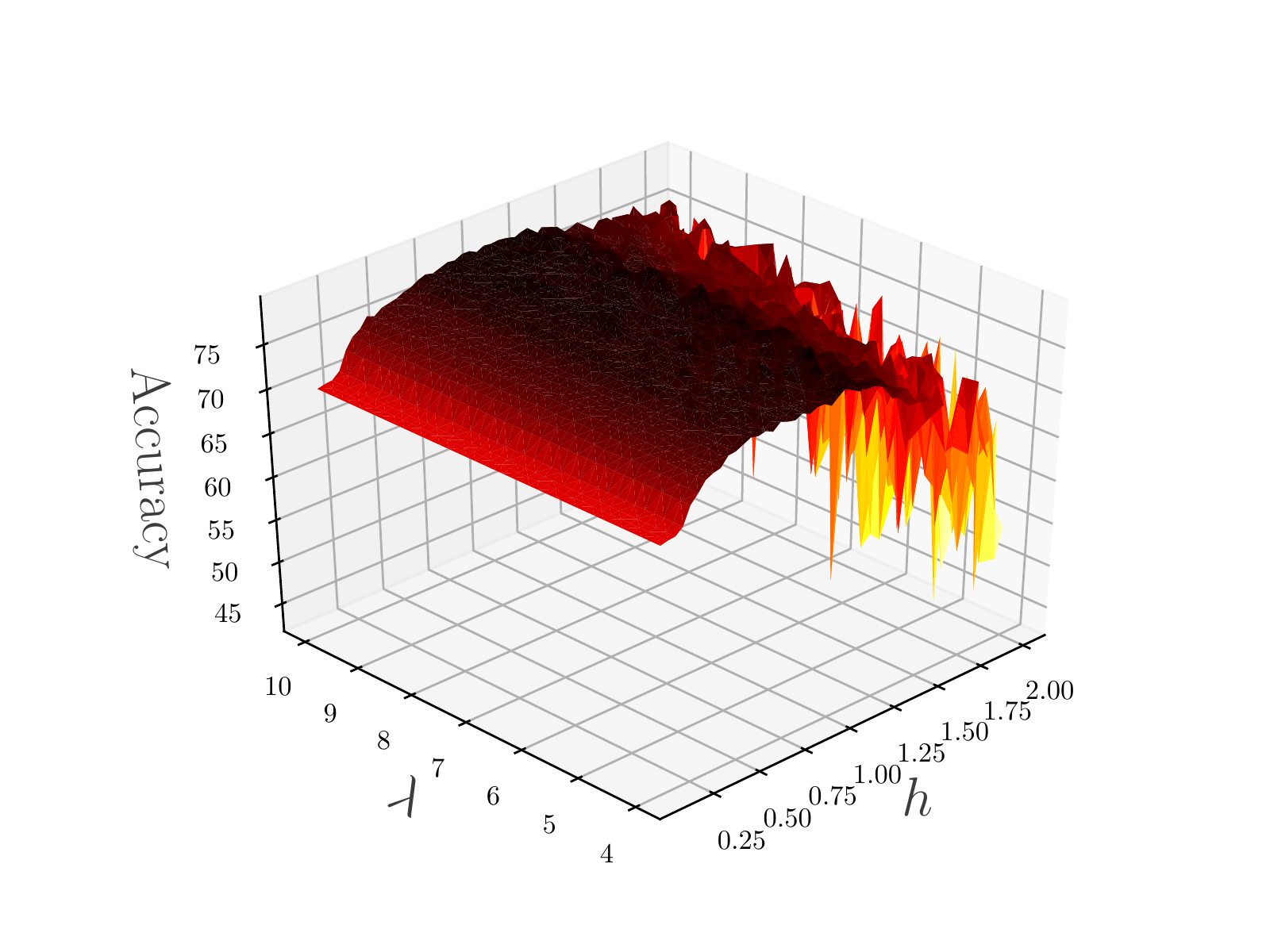}
		\caption{Grid search}
		\label{fig:gridSearch}
	\end{subfigure}%
	\begin{subfigure}{.23\textwidth}
		\includegraphics[height=1.5in]{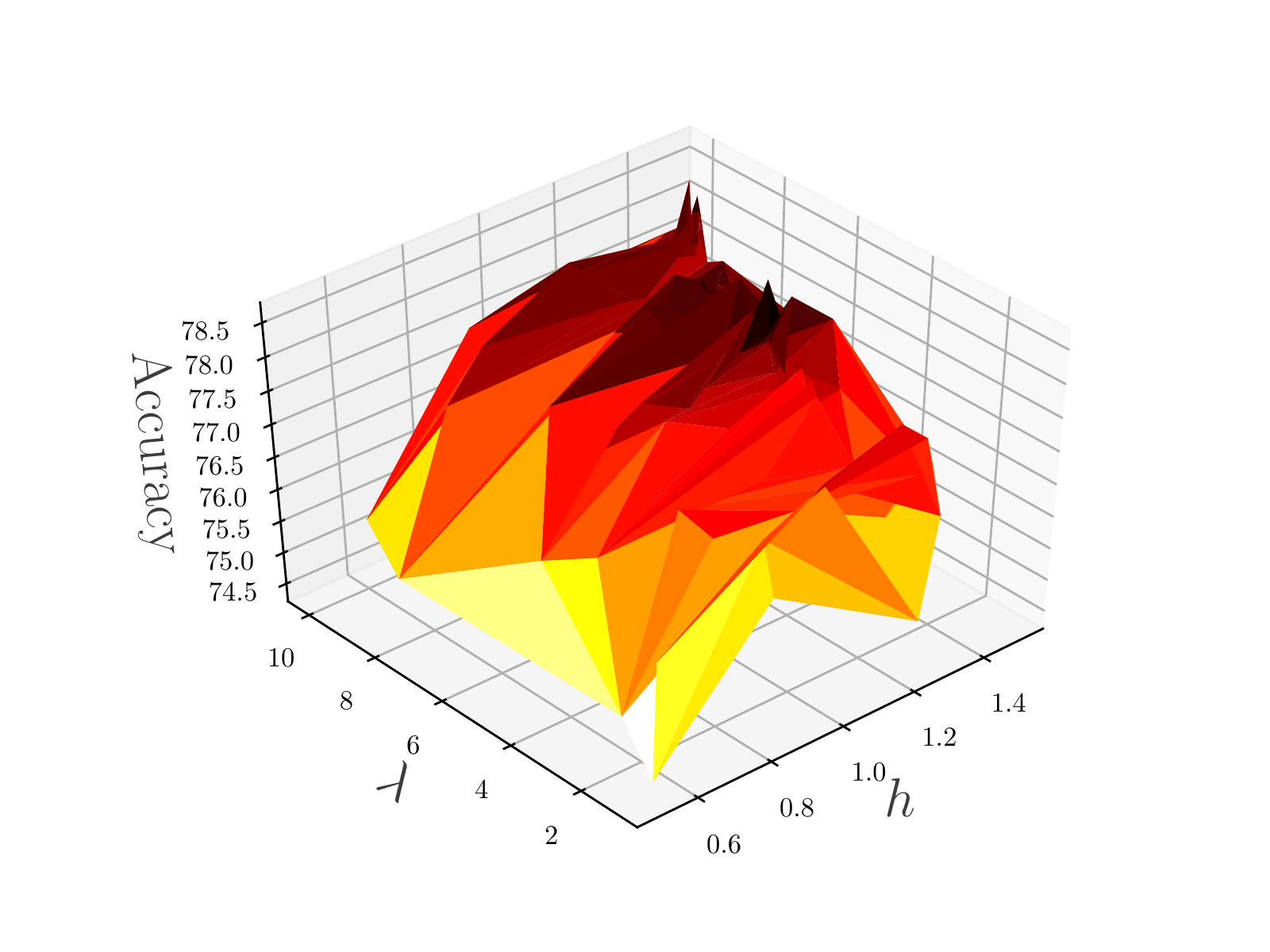}
		\caption{OpenTuner}
		\label{fig:openTuner}
	\end{subfigure}
	\caption{Left: grid search of $128^2$ runs for the SUSY dataset.
		Right: black-box optimization with 100 runs.}
\end{figure}

However, a change to $h$ requires to perform HSS reconstruction from scratch,
which is costly.
There are theoretical estimates that limit the search space for
$h$~\cite{hESTIMATE}, but it is problem dependent.
A fine grid search is too costly, see Figure \ref{fig:gridSearch}.
In contrast, the black-box optimization techniques in the OpenTuner
package~\cite{OPENTUNER} in Figure~\ref{fig:openTuner} required only
100 runs and converged to a tuning parameter with better prediction
accuracies than grid search. This technique drastically reduced the
computational requirements to select $h$ and $\lambda$.
	
\subsection{Large-Scale Prediction}

The appeal of using optimal algorithms is the ability to process large amount of data. Table~\ref{table:largeScalePrediction} shows the predictive capabilities of this method by using datasets that allow the training step to use data points in the order of millions of entries, at different dimensions. The reported prediction accuracy is tested against a \textit{test} subset of the complete dataset, that is, labeled data that were not considered during training or hyperparameter tuning.
	
\begin{table}[]
	\centering
	\small
	\begin{tabular}{|l|l|l|l|l|l|}
		\hline
		Dataset   & $N$   & $d$ & $h$ & $\lambda$ & Acc \\ \hline
		SUSY      & 4.5M  & 8   & 0.08  & 10    & 73\% \\ \hline
		MNIST     & 1.6M  & 784 & 1.1   & 10    & 99\% \\ \hline
		COVTYPE   & 0.5M  & 54  & 0.07  & 0.3   & 99\% \\ \hline
		HEPMASS   & 1.0M  & 27  & 0.7   & 0.5   & 90\% \\ \hline
	\end{tabular}
	\caption{Large-scale prediction on test data.}
	\label{table:largeScalePrediction}
\end{table}

\vspace{-10pt}
\subsection{Asymptotic Complexity}
    
\begin{figure}[H]
	\centering
	\begin{subfigure}{.23\textwidth}
		\centering
		\includegraphics[height=1.2in]{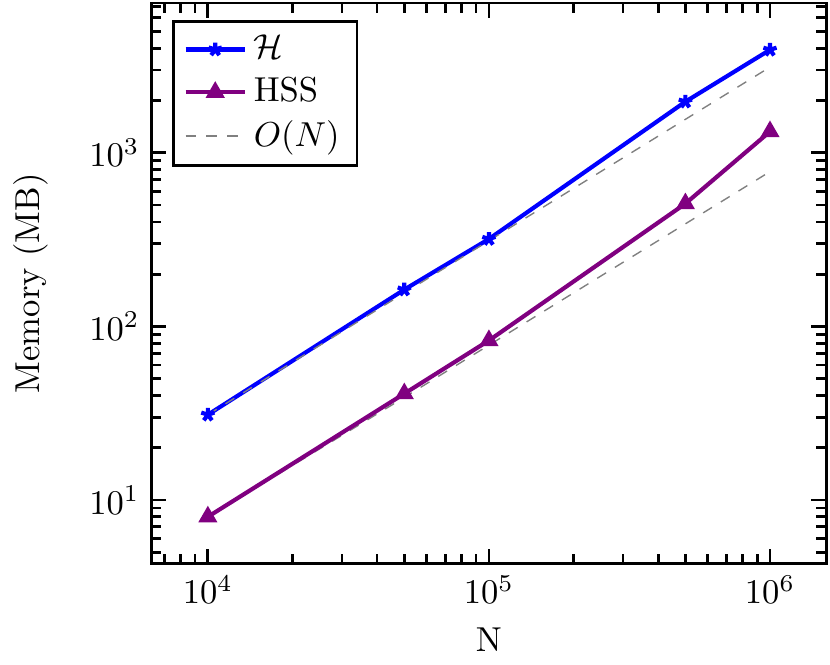}
		\caption{Memory consumption of the compressed matrices.}
		\label{fig:memoryAsymptotic}
	\end{subfigure}%
	\hspace{0.01in}
	\begin{subfigure}{.23\textwidth}
		\centering
		\includegraphics[height=1.2in]{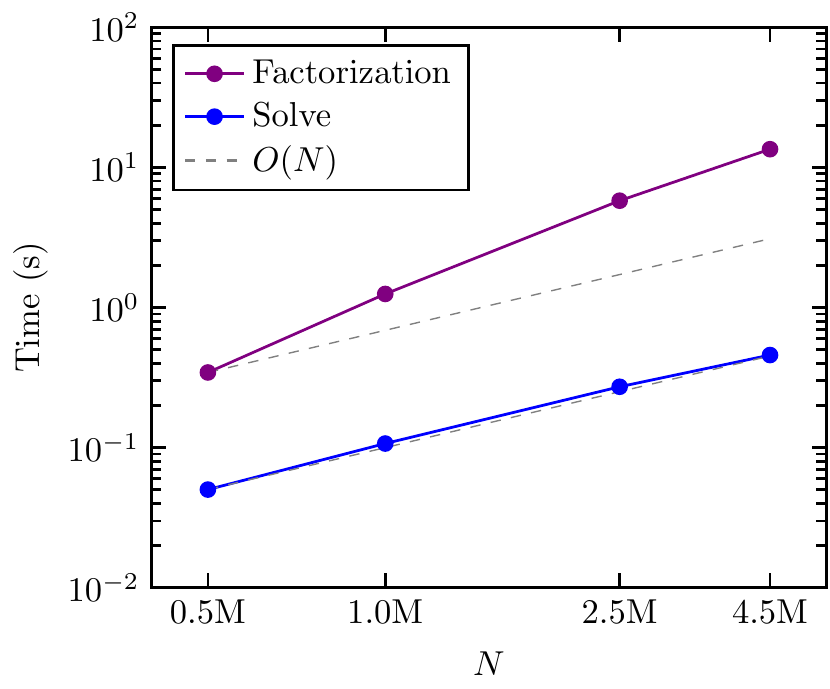}
		\caption{Time of the factorization and solve stages with HSS.}
		\label{fig:factorizationAndSolveAsymptotic}
	\end{subfigure}
	\caption{Asymptotic complexity of memory and time
 using the SUSY $N$=4.5M dataset.}
	\label{fig:hyperparameterTuning}
\end{figure}

When hierarchical matrix approximations have constant ranks, such as in a broad class of elliptic partial differential equations, the HSS memory consumption are number of operations are strict $\mathcal{O}(N)$; however, recent theoretical results show that the numerical rank of kernel matrices in high dimension depends on the dimension of the dataset \cite{RBF}. We experimentally confirm this rank growth in the additional memory requirements (Figure~\ref{fig:memoryAsymptotic}) and time for factorization (Figure~\ref{fig:factorizationAndSolveAsymptotic}).
    
The major benefit of the \textit{near}-linear complexity in factorization and memory is that this method enables the use of kernel matrices for the large datasets. As an example, storing a 1M dense matrix requires 8,000GB, whereas the HSS construction used in this work just required 1.3 GB. A similar argument can be made for the factorization of such a matrix, with which a traditional Cholesky factorization of $\mathcal{O}(n^3)$ is intractable at this scale.
    
\subsection{Performance Details}
    
Table \ref{table:performance} shows the performance of the main algorithmic steps of our method. The first step is the $\mathcal{H}$ construction to accelerate the otherwise $\mathcal{O}(n^2)$ HSS sampling. For instance, with na\"{\i}ve HSS sampling, the construction process of a kernel matrix required more than 2 hours for $n$=0.5M, now with the $\mathcal{H}$ matrix sampling we can construct an HSS matrix for $n$=4.5M in about 10 minutes.
    
Currently, our $\mathcal{H}$ matrix implementation is only a prototype 
code and is not optimized at all. Although it enabled us to experiment
with datasets as large as reported in the literature, it is only capable of
effectively using a subset of the processes that the HSS code can
use (factorization and solve). Nonetheless, the successful synergy between
the $\mathcal{H}$ and HSS matrix formats motivates our future work to develop 
a robust and more scalable distributed memory $\mathcal{H}$ matrix code. With that, we expect to achieve much faster $\mathcal{H}$ construction time and sampling time.

\begin{table}[]
	\centering
	\begin{tabular}{|l|l|l|l|l|}
		\hline
		& \multicolumn{2}{c|}{SUSY} & \multicolumn{2}{c|}{COVTYPE}           \\ \hline
		\multicolumn{1}{|c|}{Cores} & 32      & 512      & 32       & 512     \\ \hline
		$\mathcal{H}$ construction  & 173.7   & 18.3     & 36.5     & 32.2    \\ \hline
		HSS construction            & 3344.4  & 726.7    & 432.3    & 239.7   \\ \hline
		$\longrightarrow$ Sampling  & 2993.5  & 662.1    & 305.2    & 178.4   \\ \hline
		$\longrightarrow$ Other     & 350.9   & 64.6     & 127.1    & 61.3    \\ \hline
		Factorization               & 14.2    & 3.3      & 26.5     & 4.6     \\ \hline
		Solve                       & 0.5     & 0.3      & 0.5      & 0.4     \\ \hline
	\end{tabular}
	\caption{Timing details in seconds for the SUSY and COVTYPE datasets.}
	\label{table:performance}
\end{table}
	
\begin{figure}[]
	\centering\includegraphics[width=3in]{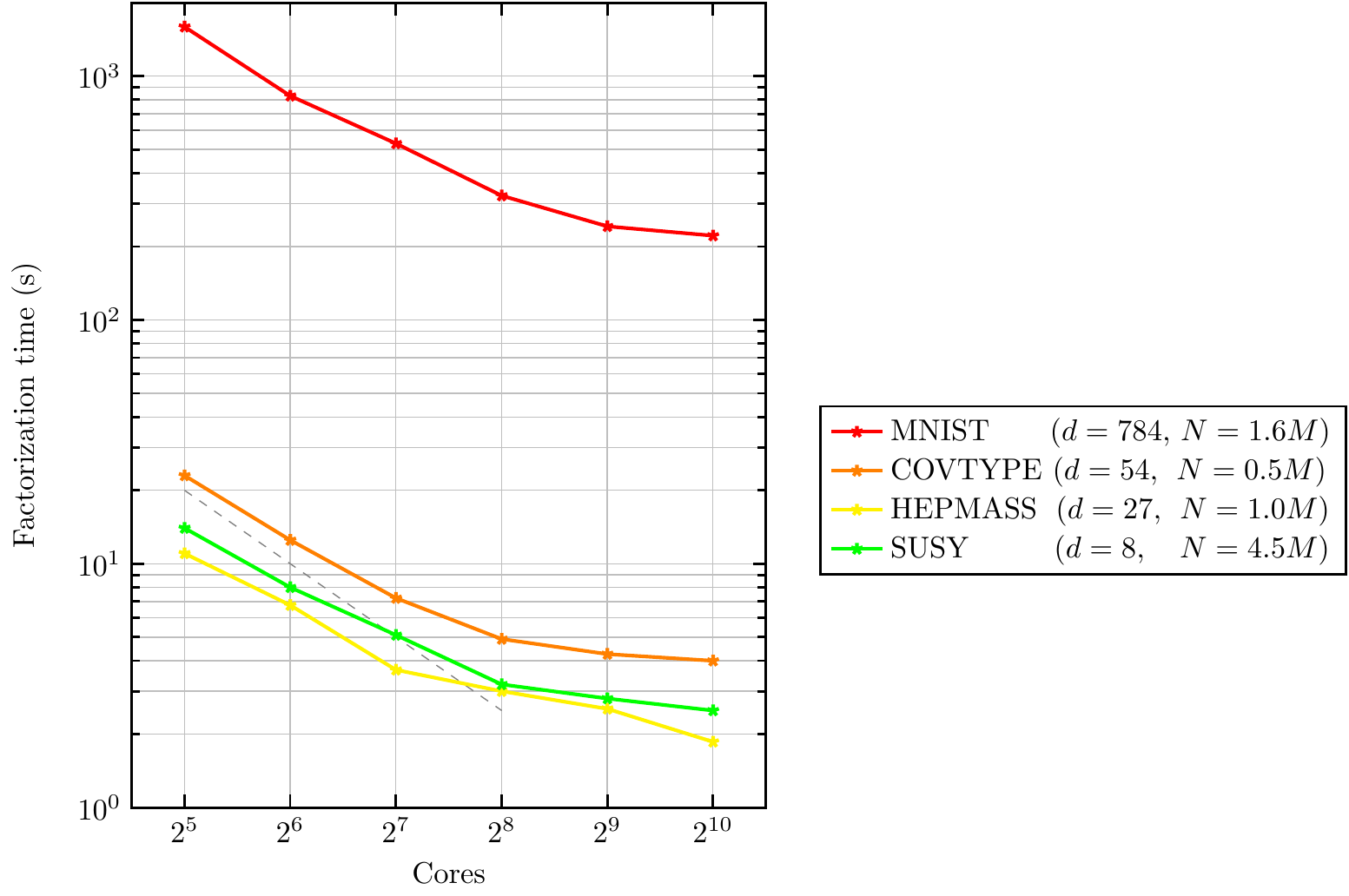}
	\caption{Strong scaling of large-scale datasets.}
	\label{fig:scaling}
\end{figure}

Figure~\ref{fig:scaling} shows a strong scaling experiment up to 1,024 cores of the factorization phase of the kernel computation for the datasets used in Table~\ref{table:performance}. As mentioned in the previous section, a scalable and fast factorization for large datasets is critical for kernel matrix computations. We refer the reader to \cite{1} for the distributed memory parallelization aspects of the ULV factorization used in this work. At large core count, the number of degrees of freedom per core decreases dramatically, while communication time starts to dominates, hence runtime starts to depart from the linear scaling line. The wall-clock time, however, is in the order of a handful of seconds even for the largest dataset considered in this work. Note that even though the SUSY dataset is larger than the MNIST dataset the overall factorization time is larger for MNIST, this is due to the fact that the dimension of MNIST is bigger than SUSY, and the dimension of the dataset has a direct effect on the necessary ranks of the approximation, and therefore in the required number of operations.

\vspace{-10px}
\section{Conclusions}

We showed that the HSS linear solvers, such as the one as implemented in STRUMPACK, are useful for a completely new area of machine learning applications. We proposed to use several relatively sophisticated ways to preprocess the kernel matrix (i.e. cluster points in the dataset), and showed that the preprocessing can significantly improve the compression rate. In all the literature we have seen the authors used either natural or k-d tree preprocessing and did not perform comparison of the various ordering techniques.
    
We presented performance data of an HSS-based complexity solver for kernel matrices scaling up to 1,024 cores of the NERSC Cori supercomputer. The construction of the HSS matrices used different preprocessing methods to minimize memory consumption for the solution of the classification task on high dimensional datasets at high prediction  accuracy.

Preliminary results show that an incomplete factorization as described in this work might be an effective preconditioner for the iterative solution of kernel matrices. We will report on the trade-offs and effectiveness of this strategy in future work.
	
\section*{Acknowledgements}
We thank Wissam Sid Lakhdar for the fruitful discussions about hyperparameter
tuning and for facilitating his OpenTuner scripts.

This research was supported by the Exascale Computing Project (17-SC-20-SC),
a collaborative effort of the U.S. Department of Energy Office of Science
and the National Nuclear Security Administration.

The research of the first author was supported in part by an appointment
with the NSF Mathematical Sciences Summer Internship Program sponsored
by the National Science Foundation, Division of Mathematical
Sciences (DMS).

\end{document}